\pdfoutput=1

\documentclass[11pt]{article}

\usepackage[preprint]{acl}

\usepackage{times}
\usepackage{latexsym}
\usepackage{multirow}

\usepackage[T1]{fontenc}

\usepackage[utf8]{inputenc}

\usepackage{microtype}

\usepackage{graphicx}
\usepackage{amsmath}        
\usepackage{booktabs}       
\usepackage{subfigure}
\usepackage{subcaption}

\title{RWKV-X: A Linear Complexity Hybrid Language Model}

\author{Haowen Hou\textsuperscript{1\textdagger} 
  \and Zhiyi Huang\textsuperscript{2\textdagger} 
  \and Kaifeng Tan\textsuperscript{3}
  \and Rongchang Lu\textsuperscript{4} 
  \and Fei Richard Yu\textsuperscript{1} \\
  \textsuperscript{1}Guangdong Laboratory of Artificial Intelligence and Digital Economy (SZ), Shenzhen, China \\
  \textsuperscript{2}College of Information Science and Engineering, Hohai University, Nanjing, China\\
  \textsuperscript{3}College of Computer Science and Software Engineering, Shenzhen University, Shenzhen, China \\
  \textsuperscript{4}School of Ecological and Environmental Engineering, Qinghai University, Xining, China \\
  \texttt{\{houhaowen, yufei\}@gml.ac.cn}
  \thanks{
  This work is supported in part by Huawei AI University Collaboration Program under Grant 43cd7230759d4951909e1ff84b171de6. \\
  \textdagger. Contributed Equally
    }
  }

\begin{document}
\maketitle
\begin{abstract}
In this paper, we introduce RWKV-X, a novel hybrid architecture that combines the efficiency of RWKV for short-range modeling with a sparse attention mechanism designed to capture long-range context. 
Unlike previous hybrid approaches that rely on full attention layers and retain quadratic complexity, RWKV-X achieves linear-time complexity in training and constant-time complexity in inference decoding.
We demonstrate that RWKV-X, when continually pretrained on 64K-token sequences, achieves near-perfect accuracy on the 64K passkey retrieval benchmark. 
It consistently outperforms prior RWKV-7 models on long-context benchmarks, while maintaining strong performance on short-context tasks. These results highlight RWKV-X as a scalable and efficient backbone for general-purpose language modeling, capable of decoding sequences up to 1 million tokens with stable speed and memory usage.
To facilitate further research and analysis, we have made the checkpoints and the associated code publicly accessible at the following GitHub repository: \href{https://github.com/howard-hou/RWKV-X}{https://github.com/howard-hou/RWKV-X}.
\end{abstract}

\section{Introduction}

\begin{figure}[!ht]
    \centering
    \subfigure[RWKV-7-4K 2.9B]{%
        \includegraphics[width=\linewidth]{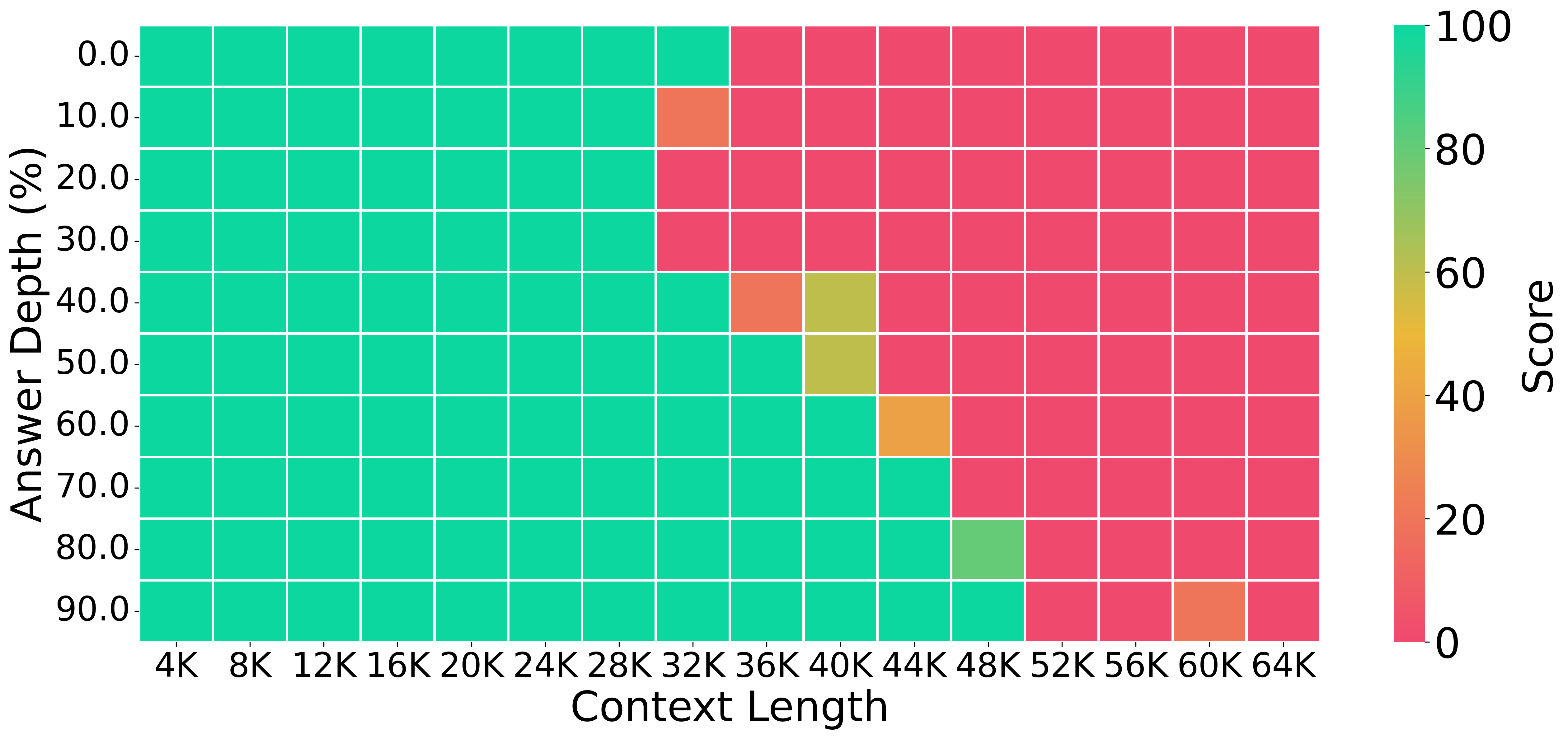}
        \label{fig:passkey_rwkv7-4k}
    }
    \subfigure[RWKV-7-128K 2.9B]{%
        \includegraphics[width=\linewidth]{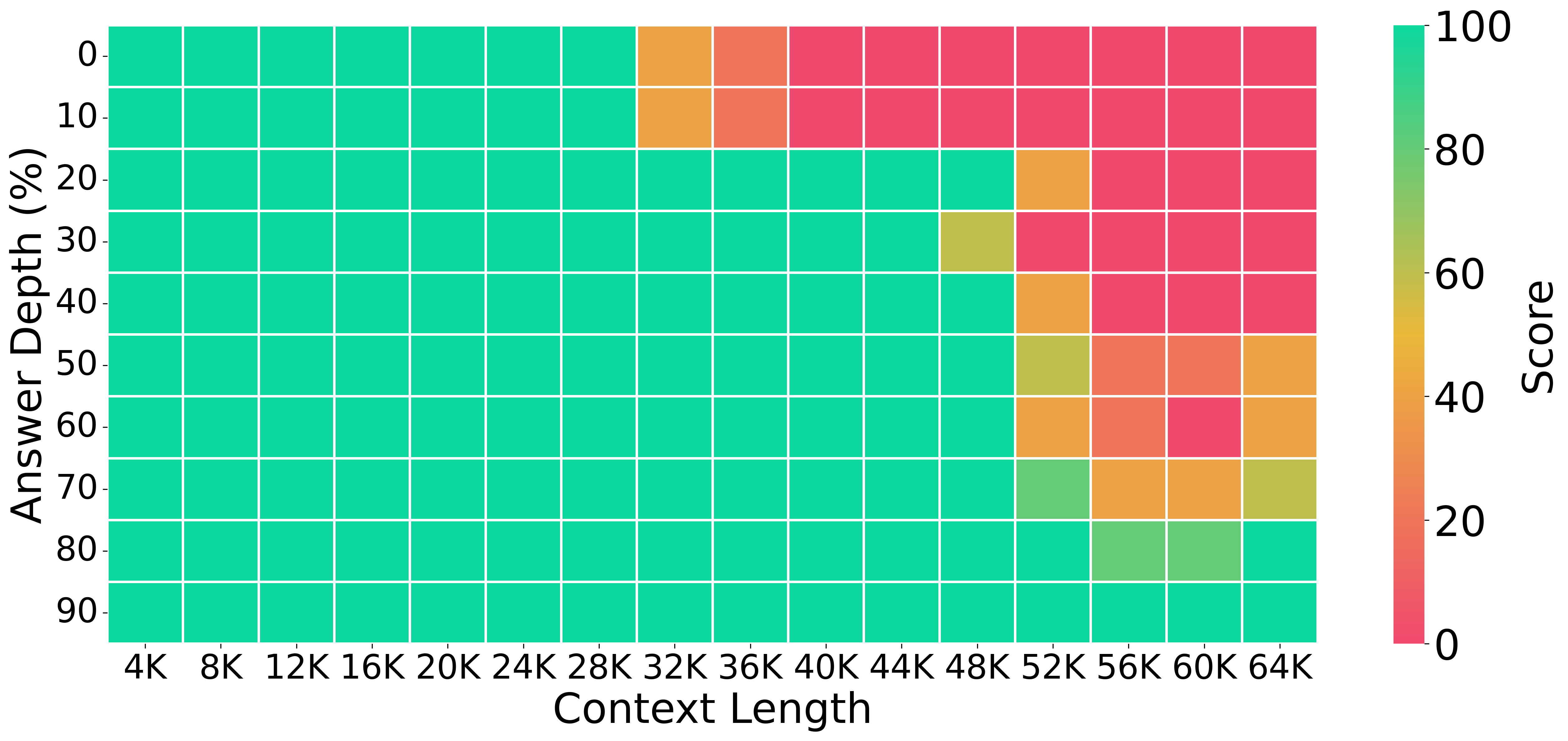}
        \label{fig:passkey_rwkv7-128k}
    }
    \subfigure[RWKV-X-64K 3.6B]{%
        \includegraphics[width=\linewidth]{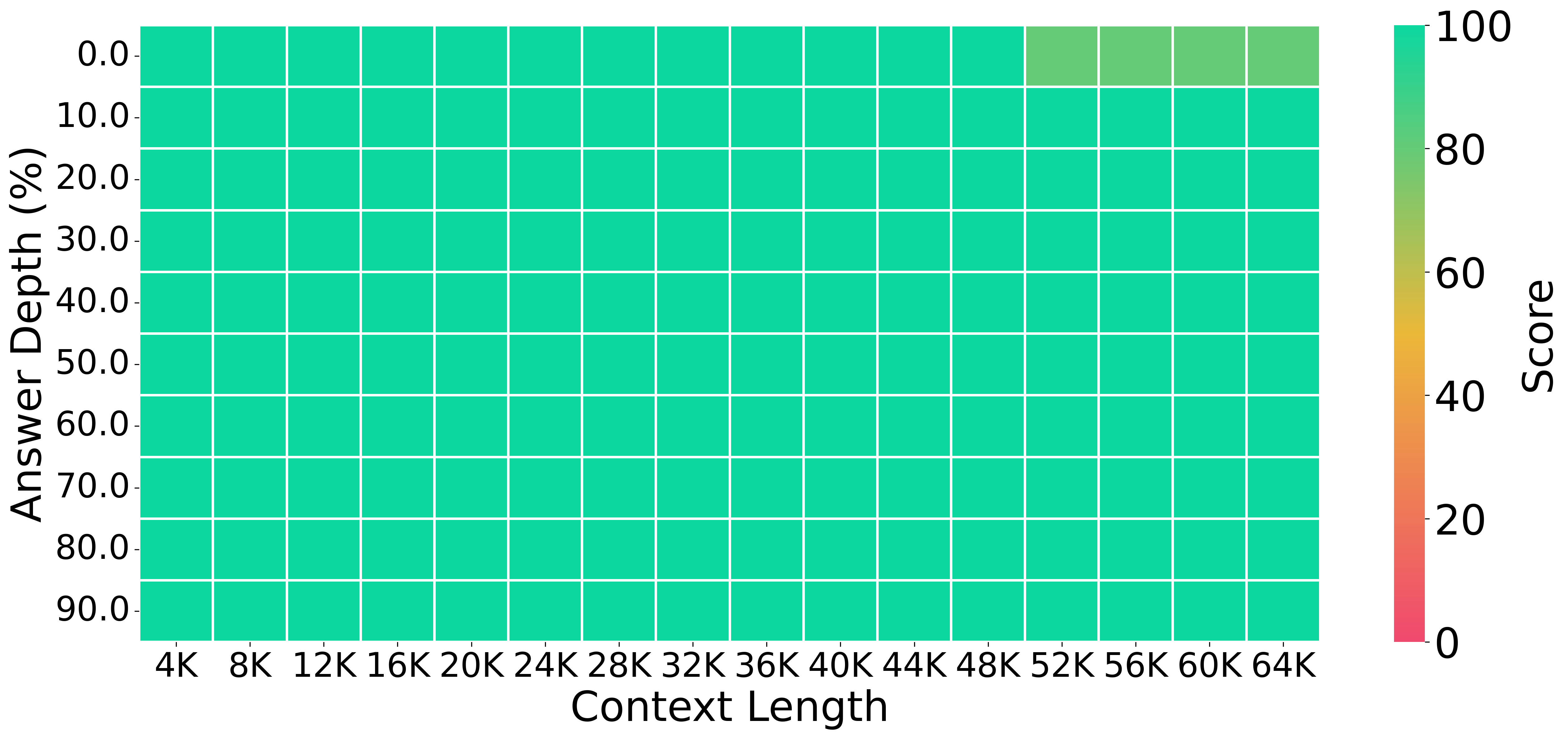}
        \label{fig:passkey_rwkvx-64k}
    }
    \caption{Passkey retrieval performance of RWKV-X models on documents up to 64K tokens. Results are shown for: (a) RWKV-7 pretrained with a 4K context length; (b) RWKV-7 after continual pretraining with a 128K context length; and (c) RWKV-X trained with continual pretraining on a 64K context length.}
    \label{fig:passkey}
\end{figure}

Transformers have become the foundation of modern large language models (LLMs), but their quadratic complexity in sequence length poses significant limitations when scaling to long-context inputs. 
To address this, a range of alternatives has emerged, including Linear Attention models~\cite{katharopoulos2020transformers}, State Space Models (SSMs) such as Mamba~\cite{gu2023mamba}, and Linear RNNs such as DeltaNet~\cite{yang2024parallelizing} and RWKV~\cite{peng2023rwkv, peng2025rwkv7gooseexpressivedynamic, peng2024eagle}.
Linear RNN-based architectures demonstrate competitive performance compared to Transformers under similar model sizes and training budgets, while significantly reducing inference costs.

However, despite their efficiency, current linear architectures still struggle with long-context understanding. 
As shown in Figure~\ref{fig:passkey_rwkv7-4k}, RWKV-7 (2.9B) achieves high accuracy on passkey retrieval up to 28K tokens, but performance rapidly degrades beyond that point. While continual pretraining with 128K-length data offers modest improvements (Figure~\ref{fig:passkey_rwkv7-128k}), long-context limitations remain. 
This limitation is not unique to RWKV; similar observations have been reported in other architectures such as Mamba~\cite{chen2024stuffed}, highlighting a broader challenge for this class of models~\cite{arora2023zoologymeasuringimprovingrecall}.

A promising approach to mitigating this limitation is the use of hybrid models that combine full attention with linear attention layers, as demonstrated in systems such as Jamba~\cite{Lieber2024JambaAH} and Zamba~\cite{glorioso2024zambacompact7bssm}. 
While these architectures improve long-context performance to some extent, their reliance on full attention layers preserves quadratic complexity, resulting in memory bottlenecks during inference over very long sequences.

In this work, we propose RWKV-X, a novel hybrid model with linear complexity that combines the strengths of RWKV for modeling short-range dependencies and sparse attention for capturing long-range context.
As shown in Figure~\ref{fig:passkey_rwkvx-64k}, RWKV-X is continually pretrained on 64K-token sequences and achieves near-perfect accuracy on the 64K passkey retrieval benchmark.

Experimental results demonstrate that RWKV-X substantially improves performance on long-context benchmarks while maintaining competitive accuracy on short-context tasks. In terms of system efficiency, RWKV-X achieves linear-time complexity ($O(N)$) during training and constant-time complexity ($O(1)$) during inference decoding. These results highlight RWKV-X as a highly effective and scalable backbone for general-purpose language modeling across both short and long contexts.

Our main contributions are as follows:
\begin{itemize}
    \item We propose RWKV-X, a novel hybrid model that achieves linear complexity in both training and inference, while effectively modeling long-range dependencies.
    \item We develop a sparse attention mechanism with linear complexity, which integrates seamlessly with the RWKV architecture to enhance long-context modeling.
    \item RWKV-X outperforms baseline models on long-context benchmarks, while maintaining strong performance on short-context tasks.
    \item RWKV-X enables ultra-long-range decoding with constant inference speed and memory usage up to 1M context length.
\end{itemize}


\section{Related Work}

\subsection{Linear Complexity Language Models}
Traditional transformer-based language models primarily rely on self-attention mechanisms with quadratic complexity, resulting in significant computational demands—especially when processing long sequences. 
In response, recent research has focused on alternative architectures that achieve linear complexity while maintaining competitive performance.
Notable examples include State-Space Models~\cite{gu2023mamba}, Linear Attention models~\cite{katharopoulos2020transformers}, and Linear RNNs~\cite{orvieto2023resurrecting}.

The RWKV model\cite{peng2023rwkv} represents a noteworthy advancement, integrating characteristics of both transformers and RNNs. By maintaining a recurrent state representation similar to RNNs, while preserving the parallelizability of transformers during training, RWKV achieves linear complexity without compromising performance. The foundational RWKV-4~\cite{peng2023rwkv} architecture demonstrated that an RNN-like model could deliver competitive performance in language modeling. RWKV-5~\cite{peng2024eagle} introduced matrix-valued states and dynamic recurrence mechanisms, further enhancing efficiency and stability. RWKV-6~\cite{peng2024eagle} improved training stability, facilitating greater scalability, while RWKV-7~\cite{peng2025rwkv7gooseexpressivedynamic} incorporated dynamic state evolution, surpassing traditional attention-based models and enabling more flexible in-context learning.
These advancements position the RWKV model as an effective and efficient linear complexity language model. As a result, it offers a promising alternative to transformers, maintaining robust language modeling capabilities while being computationally efficient.

\subsection{Hybrid Language Models}
Recent advances have shown that hybrid models outperform traditional architectures in both accuracy and efficiency. For example, Mamba~\cite{gu2023mamba} integrates retrieval-based and generative components, allowing models to leverage both pre-trained knowledge and external information sources. Building on this paradigm, recent models such as Jamba~\cite{Lieber2024JambaAH}, Zamba~\cite{glorioso2024zambacompact7bssm}, and MiniMax~\cite{minimax2025minimax01scalingfoundationmodels} further enhance hybrid architectures.

Jamba improves multilingual understanding in low-resource settings by refining the interaction between retrieval and generation. Zamba introduces a compact SSM-transformer hybrid design that balances attention capabilities with parameter efficiency. MiniMax combines mixture-of-experts (MoE) with advanced sparse attention mechanisms to scale to longer contexts.

While these hybrid approaches bring notable improvements, they still rely on full attention mechanisms, and thus \textbf{inherently preserving the $O(N^2)$ complexity}, which limits their scalability to truly long-context scenarios.

\subsection{Sparse Attention}
Native Sparse Attention~\cite{yuan2025nativesparseattentionhardwarealigned} reduces token interactions by structuring keys and values into temporal blocks and processing them through three distinct attention paths: compressed coarse-grained tokens, selectively retained fine-grained tokens, and sliding windows for local contextual information. This approach dynamically selects the most relevant tokens, optimizing the balance between global and local context while significantly reducing computational overhead.
Similarly, SeerAttention~\cite{gao2025seerattentionlearningintrinsicsparse}, inspired by the gating mechanism in Mixture-of-Experts (MoE) models, enhances efficiency by introducing learnable gating units within the attention mechanism.

Moreover, Mixture of Block Attention (MoBA)~\cite{lu2025mobamixtureblockattention} addresses the inefficiencies of traditional attention mechanisms by partitioning the input context into blocks and employing a gating mechanism to selectively route query tokens to the most relevant blocks. This design not only improves computational efficiency but also enables seamless transitions between full and sparse attention modes.
However, during autoregressive decoding, MoBA suffers from increasing memory usage as the KV cache grows with the sequence length, leading to a linear space complexity. As a result, it cannot guarantee constant memory consumption during inference, which limits its scalability for long-context generation.


\section{Method}

\begin{figure*}[htbp]
    \centering
    \includegraphics[width=\textwidth]{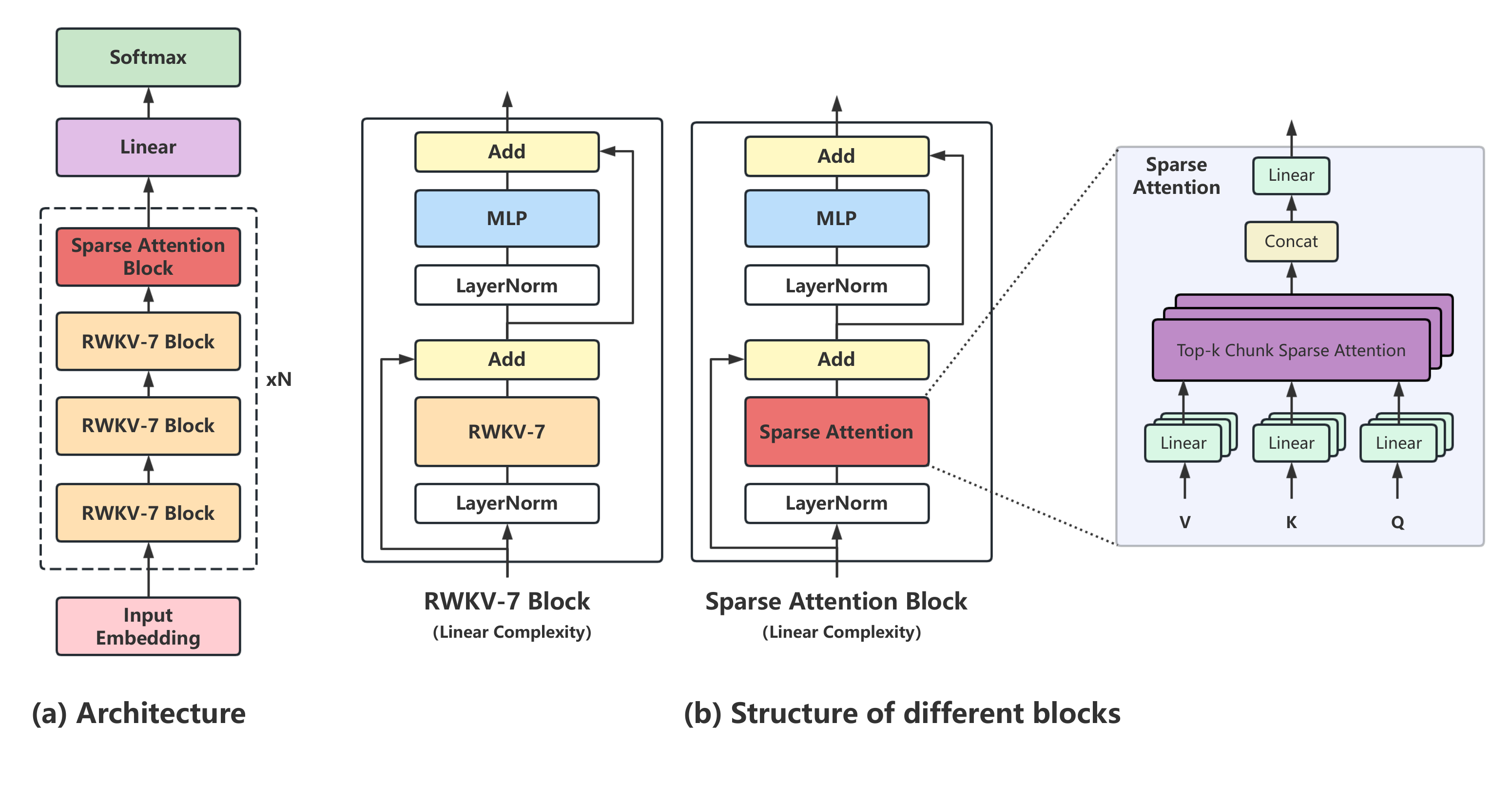}
    \caption{The architecture of RWKV-X, a hybrid model that combines RWKV-7 blocks with Sparse Attention blocks.}
    \label{fig:rwkvx_arch}
\end{figure*}

\subsection{Preliminaries}
The core of the Transformer~\cite{vaswani2017attention} architecture is the self-attention mechanism, which enables the model to compare each token with every other token in the input sequence—crucial for modeling long-range dependencies. However, for each query $q$, attention must be computed against all keys $K$ and values $V$, resulting in quadratic complexity with respect to sequence length. This makes full attention inefficient for very long sequences. The computation is defined as:
\begin{equation}
\text{Attn}(q, K, V) = \text{softmax}\left(\frac{qK^\top}{\sqrt{d_k}}\right)V
\end{equation}
Here, $d_k$ represents the dimensionality of the query and key vectors, which is used to scale the dot product to prevent extremely large values that could destabilize the softmax operation.

To overcome these limitations, recent work explores architectures that combine linear attention with dynamic state control. RWKV-7 offers an efficient alternative to Transformers for long-sequence tasks by blending the recurrence of RNNs with the parallelism of Transformers. It leverages a generalized Delta Rule~\cite{schlag2021linear} with vector-valued gating and context-dependent learning rates to enhance expressivity and efficiency. Inspired by DeltaNet, RWKV-7 further decouples state removal and addition, enabling channel-wise updates of state information.

The core mechanism of RWKV-7 introduces and optimizes the generalized Delta Rule as the foundation for state evolution. The state $S_t$ evolution and transition matrix $M_t$ are formulated as follows:
\begin{align}
    & S_t = S_{t-1}M_t + v_t^\top \cdot \tilde{k}_t \\
    & M_t =  \text{diag}(w_t) - \hat{\kappa}_t^\top (a_t \odot \hat{\kappa}_t)
\end{align}

where \(w_t\) is the data-dependent decay vector, \(a_t\) is the context-dependent learning rate, \(\hat{\kappa}_t\) is the normalized removal key, \(\tilde{k}_t\) is the replacement key, and \(v_t\) is the value vector.

In this paper, we explore the integration of Transformer and RWKV-7 architectures to build a hybrid model with linear complexity that combines the strengths of both. This model addresses the limitations of each architecture, offering a more efficient and scalable solution for sequence modeling.

\subsection{Top-\(k\) Chunk Sparse Attention}

As shown in Figure~\ref{fig:rwkvx_arch}, we introduce Top-\(k\) Chunk Sparse Attention in RWKV-X, which draws inspiration from Mixture of Block Attention (MoBA)~\cite{lu2025mobamixtureblockattention} and incorporates KV cache management to achieve constant-time complexity during inference decoding.
Instead of computing attention over the full sequence, Top-\(k\) Chunk Sparse Attention enables each query to attend only to a small, relevant subset of the input, significantly reducing computational cost.

First, given an input sequence of length $N$, it is divided into $n$ equal-sized chunks, where each chunk has a size of $B$.
For each query token $q$, the model computes a relevance score $s_i$ for each chunk $i$ using the inner product between $q$ and the mean-pooled key vectors in that chunk:

\begin{equation}
s_i = q \cdot \left( \frac{1}{B} \sum_{j=1}^{B} k_j^{(i)} \right), \quad i = 1, \dots, n
\end{equation}
where \( k_j^{(i)} \) denotes the \( j \)-th key vector within the \( i \)-th chunk.

Next, the indices of the top-\( k \) chunks with the highest scores are selected:
\begin{equation}
\mathcal{I} = \text{TopK}\left( \{ s_i \}_{i=1}^{n}, k \right)
\end{equation}
where \( \mathcal{I} \subseteq \{1, \dots, n\} \) is the set of selected chunk indices.

Finally, attention is computed only over the key-value pairs from the selected chunks:
\begin{equation}
\text{Attn}(q, K_{\mathcal{I}}, V_{\mathcal{I}}) = \text{softmax}\left( \frac{q K_{\mathcal{I}}^\top}{\sqrt{d_k}} \right) V_{\mathcal{I}}
\end{equation}

Top-\(k\) Chunk Sparse Attention reduces the quadratic complexity of standard attention by restricting each query to attend only to a small set of highly relevant chunks. As a result, it preserves the model’s ability to capture long-range dependencies while significantly improving efficiency, particularly on long-context sequences.

\subsubsection{KV Cache Management}
In the decoding stage of inference, without KV cache management, the sequence length continuously increases. 
Under such conditions, applying Top-\(k\) Chunk Sparse Attention alone cannot maintain constant space complexity. 

To address this, we propose a KV cache management mechanism inspired by SnapKV~\cite{li2024snapkvllmknowslooking}, aiming to maintain a constant-size cache during decoding. 

We first split the past cache into earlier cached states \((K_{\text{past}}, V_{\text{past}})\) and the recent observation window \((K_{\text{obs}}, V_{\text{obs}})\).

We then compute an importance score vector \(C\) over \(K_{\text{past}}\) by summing the softmax-normalized attention scores between \(Q_{\text{obs}}\) and \(K_{\text{past}}\):

\begin{equation}
C = \sum_{i=1} \text{softmax}\left( \frac{Q_{\text{obs}} K_{\text{past}}^\top}{\sqrt{d_k}} \right)[i, :]
\label{eq:importance_score_short}
\end{equation}

The top-\(m\) keys and values are selected based on \(C\), where \(m\) is a predefined memory budget.

Finally, we reconstruct the compressed past cache by concatenating the selected entries with the observation window, ensuring constant memory usage without loss of critical information. 
We refer the reader to Appendix~\ref{sec:kv_cache} for more details on KV cache management.

\subsubsection{Complexity Analysis}
\textbf{Complexity Analysis during Training.}
The complexity of Top-$k$ Chunk Sparse Attention is \( O(kBN) \), where \(N\) is the sequence length, \(B\) is the chunk size, and \(k\) is the number of selected chunks. 
Since \(k\) and \(B\) are small constants, the complexity scales linearly with \(N\), approaching \(O(N)\).

\textbf{Complexity Analysis during Decoding.}
During decoding, the past KV cache is compressed to a fixed size \(m\).
Each step computes attention over \(O(kB + L_{\text{obs}})\) entries, where \(L_{\text{obs}}\) is the observation window size.
As \(k\), \(B\), and \(L_{\text{obs}}\) are small constants, the total decoding complexity over \(N\) tokens remains \(O(N)\).

We summarize the computational complexity and memory usage of different LLM architectures in Table~\ref{tab:complexity_comparison}, covering both training and decoding scenarios.

\begin{table}[h]
\centering
\caption{Comparison of computational complexity and memory usage among different LLM architectures, in terms of training complexity per sequence and decoding complexity per token.}
\label{tab:complexity_comparison}
\resizebox{\columnwidth}{!}{
\begin{tabular}{lccc}
\toprule
\textbf{Method} & \textbf{Training} & \textbf{Decoding} & \textbf{Memory} \\
 & \textbf{Complexity} & \textbf{Complexity} & \textbf{Usage} \\
\midrule
Full Attention & \(O(N^2)\) & \(O(N)\) & \(O(N)\) \\
RWKV-7 & \(O(N)\) & \(O(1)\) & \(O(1)\) \\
Top-$k$ Chunk & \(O(kBN)\) & \(O(1)\) & \(O(1)\) \\
RWKV-X & \(O(kBN+N)\) & \(O(1)\) & \(O(1)\) \\
\bottomrule
\end{tabular}
}
\end{table}

\subsection{RWKV-X}
As shown in Figure~\ref{fig:rwkvx_arch}, RWKV-X is a hybrid architecture that combines RWKV-7 blocks~\cite{peng2025rwkv7gooseexpressivedynamic} with sparse attention blocks.  
To enhance the model’s capacity for modeling long-range dependencies, sparse attention blocks are periodically inserted between RWKV blocks.

\subsubsection{Block Expansion Method}

RWKV-X is not trained from scratch. Instead, it draws inspiration from the block expansion method of LLaMA Pro~\cite{wu-etal-2024-llama}, adopting interleaved block expansion and a zero-initialization mechanism. 
This approach minimizes the number of parameters requiring random initialization, ensuring compatibility with the previous RWKV-7 model. 
Subsequently, RWKV-X undergoes a two-stage training process for block expansion.
In the first stage, short texts with a context length of 1024 from the MiniPile dataset~\cite{kaddour2023minipilechallengedataefficientlanguage} are used to train the model. 
During this training phase, all parameters except those of the newly added blocks are frozen. 
This approach brings the parameters to an aligned state, laying a solid foundation for subsequent long-context pretraining.
The specific techniques used for long-context continual pretraining are introduced in the following subsection.

\subsubsection{Long-context Continual Pretraining}

In the second stage of long-context continual pretraining, we utilize the ProLong-64K training dataset~\cite{gao2025trainlongcontextlanguagemodels}. Training is conducted with a context length of 64K tokens, and the total training volume amounts to 1 billion tokens.
During this phase, all model parameters—including those previously frozen—are unfrozen and jointly optimized.

In the long-context continual pretraining stage, we employ the Long-context Cross-Entropy (LongCE) loss~\cite{fang2025wrongperplexitylongcontextlanguage} to emphasize critical tokens. 
The LongCE loss assigns dynamic weights to each token based on the traditional cross-entropy loss. 
Critical tokens receive weights greater than 1, while ordinary or out-of-scope tokens receive weights close to 1. 
This mechanism enables the model to automatically focus on tokens with long-range contextual dependencies, enhancing its performance on long-context data.

\section{Experiments}

\subsection{Experiment Setup}
The training process of RWKV-X consists of two stages: alignment pretraining and long-context continual pretraining. In the alignment pretraining stage, the RWKV-7 blocks are frozen, with only the Sparse Attention blocks being updated. During the long context continual pretraining stage, we finetune all parameters. Details of training data and hyper-parameters can be found in Appendix \ref{sec:data_Hyperparameters}.

\subsection{Long Context Evaluation}
To better understand the long-context capabilities of RWKV-X, we conduct a case study using the Single Needle-In-A-Haystack (S-NIAH) benchmark suite from RULER~\citep{hsieh2024rulerwhatsrealcontext}, where a key-value pair is embedded within a long context and the model is required to retrieve the value when given the corresponding key. 
The results are summarized in Table~\ref{tab:S-NIAH}, with some values sourced from the Gated DeltaNet paper~\cite{yang2024gated}.

\begin{table*}[h]
\centering
\caption{Zero-shot Performance Comparison on the S-NIAH Benchmark: S-NIAH-1 (Pass-key Retrieval), S-NIAH-2 (Number in Haystack), and S-NIAH-3 (UUID in Haystack).}
\label{tab:S-NIAH}
\resizebox{\textwidth}{!}
{
\begin{tabular}{l|cccc|cccc|cccc}
\toprule
\textbf{Model} & \multicolumn{4}{c|}{\textbf{S-NIAH-1}} & \multicolumn{4}{c|}{\textbf{S-NIAH-2}} & \multicolumn{4}{c}{\textbf{S-NIAH-3}} \\
& 1K & 2K & 4K & 8K & 1K & 2K & 4K & 8K & 1K & 2K & 4K & 8K \\
\midrule
RWKV-7-0.19B & 100 & 100 & 100 & 100 & 100 & 98.4 & 5.2 & 3.2 & 98.4 & 96.8 & 27.6 & 6.6 \\
RWKV-X-0.22B & 100 & 100 & 100 & 100 & 100 & 99.6 & 18.8 & 3.6 & 99.0 & 92.6 & 40.2 & 1.6 \\
\midrule
DeltaNet-1.3B & 97.4 & 96.8 & 99.0 & 98.8 & 98.4 & 45.6 & 18.6 & 14.4 & 85.2 & 47.0 & 22.4 & - \\
Mamba2-1.3B & 99.2 & 98.8 & 65.4 & 30.4 & 99.4 & 98.8 & 56.2 & 17.0 & 64.4 & 47.6 & 4.6 & - \\
Gated DeltaNet-1.3B & 98.4 & 88.4 & 91.4 & 91.8 & 100.0 & 99.8 & 92.2 & 29.6 & 86.6 & 84.2 & 27.6 & - \\
RWKV-6-1.6B & - & - & 98.0 & - & - & - & 53.0 & - & - & - & 55.0 & - \\
\midrule
RWKV-6-3B & - & - & 100 & - & - & - & 88.0 & - & - & - & 79.0 & - \\
RWKV-7-2.9B & - & - & 100 & - & - & - & 88.0 & - & - & - & 79.0 & - \\
\textbf{RWKV-X-3.6B} & \textbf{100} & \textbf{100} & \textbf{100} & \textbf{100} & \textbf{100} & \textbf{100} & \textbf{100} & \textbf{99.8} & \textbf{100} & \textbf{100} & \textbf{99.8} & \textbf{95.6} \\
\bottomrule
\end{tabular}
}
\end{table*}

\subsection{Short Context Evaluation}

Table~\ref{tab:short_context_benchmarks} presents the performance of RWKV-X across a range of short-context language understanding benchmarks, in comparison with prior RWKV variants and several strong baselines of similar or larger scale.
Among small-scale models, RWKV-X (0.22B) achieves an average score of 51.0, which is competitive with RWKV-7 (51.8).
In the large-scale regime, RWKV-X (3.6B) achieves an average score of 71.9, closely matching the performance of RWKV-7 (2.9B, 72.8) and Qwen2.5-3B (71.4), while outperforming LLaMA3.2-3B (69.7). 
These results demonstrate that RWKV-X maintains high performance on short-context tasks while offering architectural benefits for long-context modeling, confirming its effectiveness as a general-purpose LLM backbone.

\begin{table*}[htbp]
\centering
\caption{Evaluation results on short-context benchmarks, including LAMBADA, HellaSwag, PIQA, ARC-Easy (arcE), ARC-Challenge (arcC), Winogrande, SciQ, and MMLU.}
\label{tab:short_context_benchmarks}
\resizebox{\textwidth}{!}{%
\begin{tabular}{lrcccccccccc}
\toprule
\textbf{Model} & \textbf{Tokens} & \textbf{LAMBADA} & \textbf{HellaSwag} & \textbf{PIQA} & \textbf{arcE} & \textbf{arcC} & \textbf{Winogrande} & \textbf{SciQ} & \textbf{MMLU} & \textbf{avg} \\
(Name) & (T) & acc$\uparrow$ & acc\_n$\uparrow$ & acc$\uparrow$ & acc$\uparrow$ & acc$\uparrow$ & acc$\uparrow$ & acc$\uparrow$ & acc$\uparrow$ & acc$\uparrow$ \\
\midrule
RWKV-5-0.19B & 0.6 & 38.4 & 31.9 & 61.4 & 44.2 & 19.9 & 52.9 & 76.3 & 23.1 & 43.5 \\
SmoLLM2-135M & 2.0 & 42.9 & 43.1 & 68.1 & 64.4 & 28.1 & 53.4 & 80.1 & 25.8 & 50.7 \\
RWKV-7-0.19B & 1.6 & 48.1 & 42.1 & 67.3 & 59.3 & 25.5 & 56.0 & 86.3 & 30.1 & 51.8 \\
RWKV-X 0.22B & 1.6 & 47.0 & 42.1 & 67.9 & 56.9 & 29.4 & 52.6 & 86.1 & 26.0 & 51.0 \\
\midrule
RWKV-6-3B & 2.5 & 71.7 & 68.4 & 76.4 & 71.2 & 35.6 & 66.3 & 92.2 & 28.3 & 63.8 \\
Llama3.2-3B & 15.0 & 70.5 & 73.6 & 76.7 & 74.5 & 42.2 & 68.2 & 95.6 & 56.5 & 69.7 \\
Qwen2.5-3B & 18.0 & 67.1 & 73.5 & 77.4 & 77.1 & 45.0 & 68.8 & 96.2 & 65.7 & 71.4 \\
RWKV-7-2.9B & 5.6 & 73.4 & 76.4 & 79.7 & 81.0 & 48.7 & 72.8 & 95.0 & 55.0 & 72.8 \\
RWKV-X 3.6B & 5.6 & 73.1 & 74.5 & 79.4 & 80.0 & 50.6 & 70.6 & 95.0 & 52.3 & 71.9 \\
\bottomrule
\end{tabular}%
}
\end{table*}

\begin{figure}[tbp]
    \centering
    \includegraphics[width=\columnwidth]{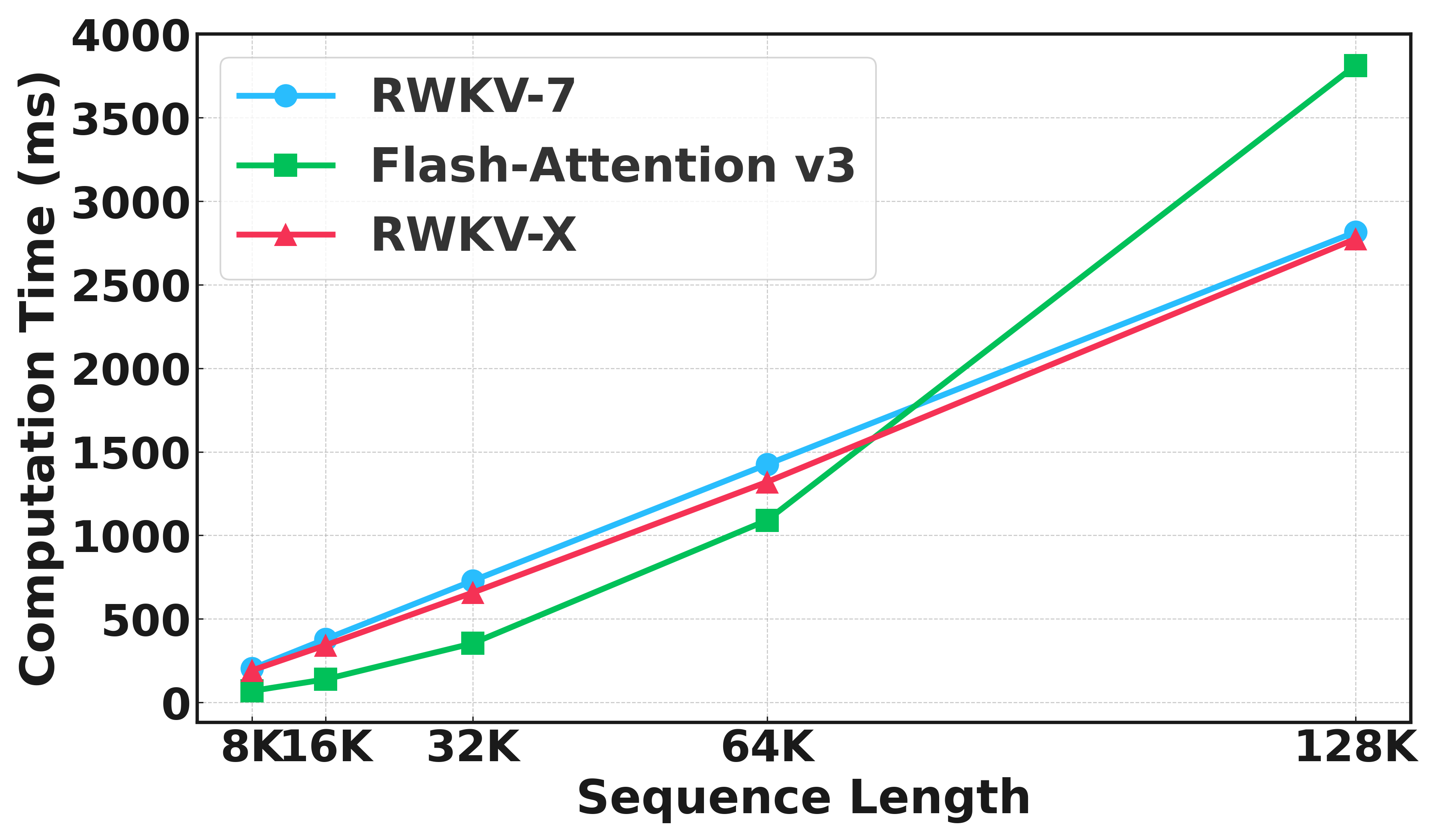}
    \caption{Prefill latency comparison between RWKV-X and a full-attention Transformer.}
    \label{fig:efficiency_comparison}
\end{figure}

\subsection{Efficiency Analysis}

Figure ~\ref{fig:efficiency_comparison} presents a comparison of prefill latency across different sequence lengths for RWKV-X, RWKV-7, and a full-attention Transformer using Flash-Attention v3~\cite{Shah2024FlashAttention3FA}. 
While Flash-Attention v3 demonstrates the lowest latency at shorter sequence lengths (8K–16K), its computation time increases steeply for longer inputs due to its quadratic complexity. 
At 128K, RWKV-X achieves a 1.37× speedup over Flash-Attention v3. Moreover, this speedup is expected to further improve as the input context length increases. 
RWKV-X exhibits near-linear scaling and consistently outperforms or matches RWKV-7, highlighting its superior efficiency and scalability for long-context inference.

\begin{figure}[tbp]
    \centering
    \includegraphics[width=\columnwidth]{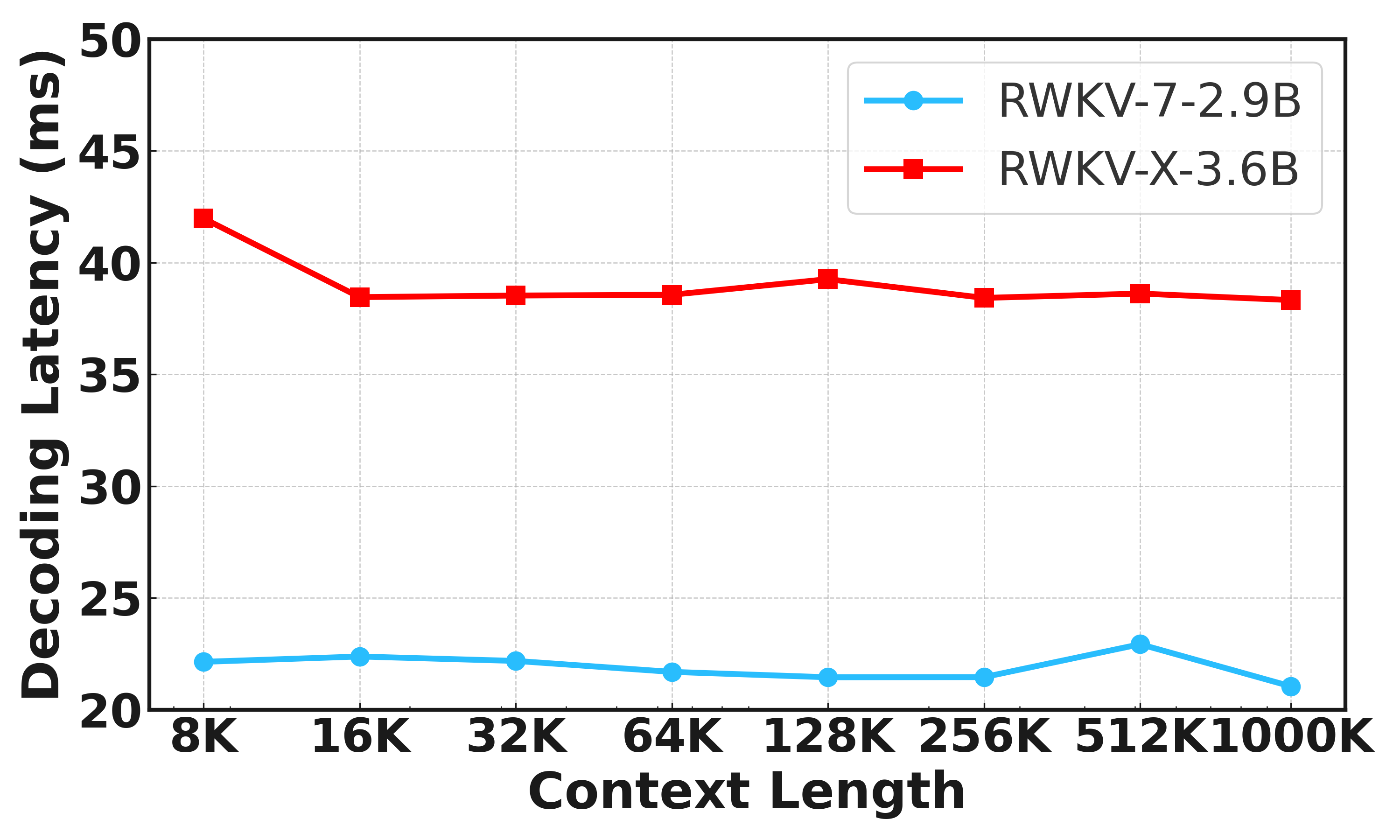}
    \caption{Decoding latency comparison between RWKV-7-2.9B and RWKV-X-3.6B models.
    The horizontal axis represents the context length (log scale).}
    \label{fig:rwkvx-latency}
\end{figure}

The results in Figure~\ref{fig:rwkvx-latency} highlight the efficiency of RWKV-X-3.6B in handling long-context decoding. 
Despite RWKV-7-2.9B achieving lower absolute latency, RWKV-X-3.6B demonstrates remarkable stability across increasing context lengths, maintaining consistent decoding times even at 1M tokens. 
RWKV-X-3.6B employs a fixed 64K KV cache, while RWKV-7-2.9B operates without additional caching mechanisms.
RWKV-X demonstrates stable decoding latency as the context length increases.

\subsection{Ablation Study}

\subsubsection{Ablation on Long context Cross Entropy}

We conduct an ablation study to evaluate the impact of the Long-context Cross-Entropy~(LongCE) loss on the performance of RWKV-X-3.6B across three tasks in the S-NIAH benchmark. As shown in Table~\ref{tab:longce_ablation}, incorporating LongCE consistently improves model performance, particularly on tasks that require longer context understanding.

On S-NIAH-1, where the task involves relatively short-context retrieval, both versions of the model— with and without LongCE—achieve perfect accuracy across all context lengths, indicating that LongCE has minimal impact when the contextual challenge is low.

However, for S-NIAH-2 and S-NIAH-3, which demand deeper reasoning over longer input sequences, the benefits of LongCE become evident. At 8K context length, the model without LongCE shows a steep drop in performance—falling to 67.0 on S-NIAH-2 and 62.6 on S-NIAH-3. In contrast, the full model with LongCE maintains high accuracy at 99.8 and 95.6, respectively. These results demonstrate that LongCE plays a crucial role in helping the model focus on semantically important tokens over extended contexts, thereby preserving performance as sequence length increases.

Overall, LongCE significantly enhances the long-context generalization ability of RWKV-X, especially in tasks where key information is sparsely distributed across the input.

\begin{table}[h]
\centering
\caption{Ablation Study on LongCE Loss using the S-NIAH Benchmark (Higher is Better).}
\label{tab:longce_ablation}
\resizebox{\linewidth}{!}{
\begin{tabular}{llcccc}
\toprule
\textbf{Model} & \textbf{Task} & \textbf{1K} & \textbf{2K} & \textbf{4K} & \textbf{8K} \\
\midrule
RWKV-X-3.6B      & S-NIAH-1 & 100.0 & 100.0 & 100.0 & 100.0 \\
w/o LongCE       & S-NIAH-1 & 100.0 & 100.0 & 100.0 & 100.0 \\
\midrule
RWKV-X-3.6B      & S-NIAH-2 & 100.0 & 100.0 & 100.0 & \textbf{99.8} \\
w/o LongCE       & S-NIAH-2 & 100.0 & 100.0 & 98.4  & 67.0 \\
\midrule
RWKV-X-3.6B      & S-NIAH-3 & 100.0 & 100.0 & 99.8  & \textbf{95.6} \\
w/o LongCE       & S-NIAH-3 & 100.0 & 100.0 & 98.4  & 62.6 \\
\bottomrule
\end{tabular}
}
\end{table}

\subsubsection{Ablation on Percentage of Attention Layers}

\begin{figure}[htbp]
    \centering
    \includegraphics[width=\columnwidth]{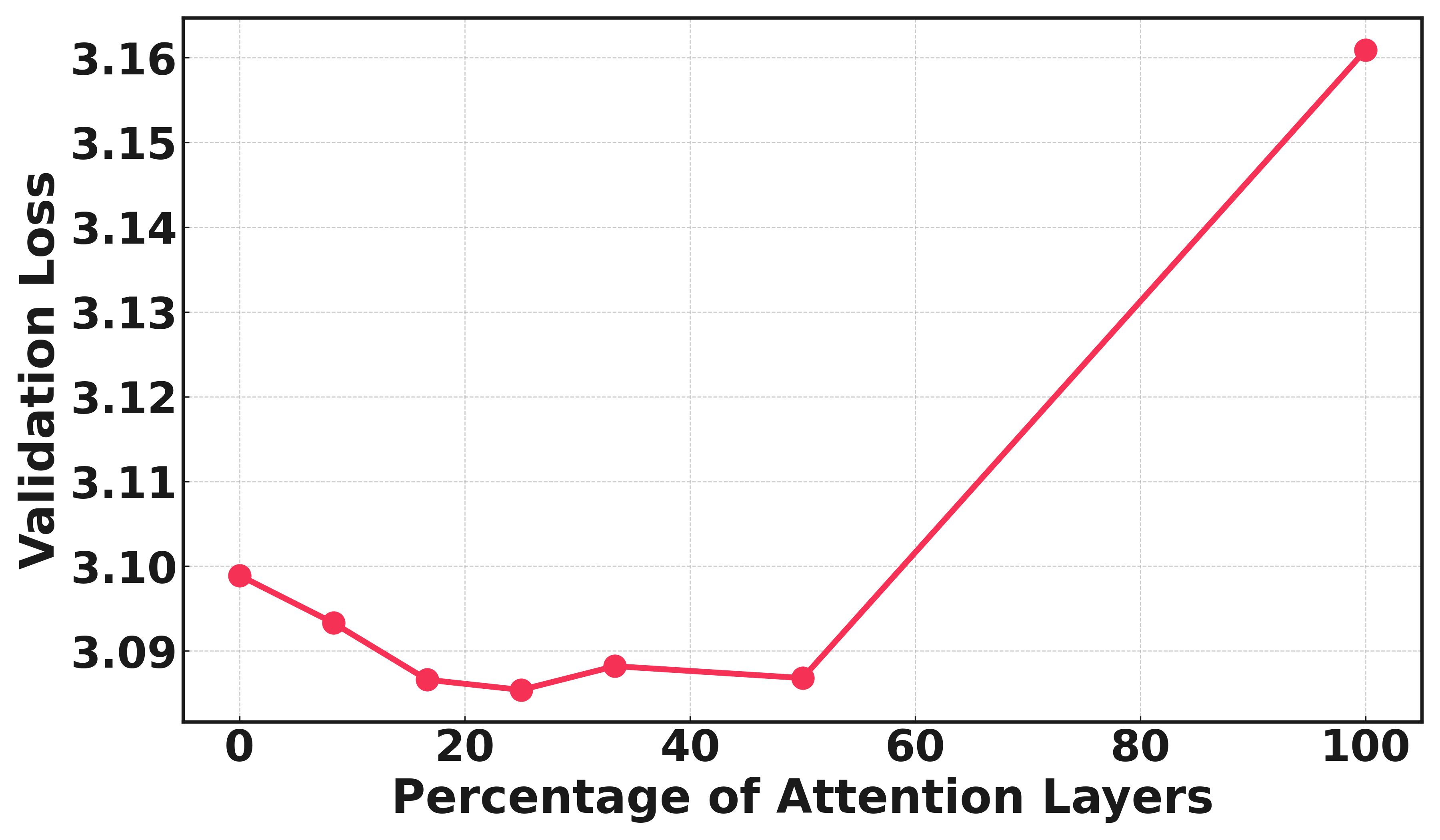}
    \caption{Validation loss vs. percentage of attention layers for 124M-parameter RWKV-X models (12 layers). 0\% = RWKV-7, 100\% = Fully Sparse-Attention Transformer.}
    \label{fig:val_loss_vs_attention_percentage}
\end{figure}

We begin by investigating how the number of Sparse Attention layers affects model performance in RWKV-X. 
In this study, we train 126M-parameter hybrid models with 12 total layers, varying the proportion of Sparse Attention layers while distributing them evenly throughout the model. Figure~\ref{fig:val_loss_vs_attention_percentage} reports the validation loss as a function of the attention layer ratio, where 0\% corresponds to RWKV-7 and 100\% corresponds to a fully Sparse-Attention Transformer.
Our results show that validation loss is minimized when approximately 25\% of the layers are Sparse Attention layers. 
This suggests that a hybrid architecture offers advantages over both the pure RWKV-7 and the fully attention-based Transformer in terms of loss.

\subsubsection{Ablation on Model Size}

Table~\ref{tab:perplexity_tokens} presents the validation loss comparison between GPT-2~\footnote{Code: https://github.com/xforcevesa/mixed-nanogpt} and RWKV-X models trained on 10 billion tokens across three model sizes. 
At each scale—small, medium, and large—RWKV-X consistently achieves lower validation loss than GPT-2. 
Notably, the performance gap becomes more pronounced as model size increases, with RWKV-X (786M parameters) outperforming GPT-2 (774M parameters) by 0.16 in validation loss at the large scale. 
These results suggest that RWKV-X exhibits better scaling behavior with model size and more effective utilization of model capacity compared to GPT-2 under the same data budget.

\begin{table}[htbp]
    \centering
    \caption{
    Validation loss comparison of GPT-2 and RWKV-X (trained on 10B tokens) across different model sizes. 
    For GPT-2, the small, medium, and large models have 124M, 350M, and 774M parameters.
    For RWKV-X, the corresponding sizes are 126M, 355M, and 786M.
    }
    \label{tab:perplexity_tokens}
    \resizebox{\columnwidth}{!}{%
        \begin{tabular}{lccccc}
            \toprule
            \textbf{Model} & \textbf{Tokens} & \textbf{Small} & \textbf{Medium} & \textbf{Large} \\
            \midrule
            GPT-2   & 10B & 3.12 & 2.84 & 2.76 \\
            RWKV-X  & 10B & 3.08 & 2.73 & 2.60 \\
            \bottomrule
        \end{tabular}
    }
\end{table}

\subsubsection{Ablation on Positional Encoding}

The results in Table~\ref{tab:pos_encoding} indicate that the choice of positional encoding has minimal impact on the validation loss of RWKV-X.
Surprisingly, the model without any positional encoding ("No Pos") slightly outperforms those using absolute positional embeddings and rotary position encoding (ROPE).
This suggests that the RNN-style recurrence mechanism inherent to RWKV-X already provides sufficient implicit positional information.
As a result, the addition of explicit positional encodings does not appear to bring additional benefit in this architecture.
In line with recent findings~\cite{Lieber2024JambaAH} and supported by our experimental results, we elect to exclude positional encodings from the RWKV-X models.

\begin{table}[htbp]
    \centering
    \caption{
    Validation loss of RWKV-X under different positional encoding schemes. 
    “No Pos” indicates no positional encoding; “Abs Pos” uses absolute positional encoding; 
    “ROPE” applies rotary position encoding.
    }
    \label{tab:pos_encoding}
    \resizebox{\columnwidth}{!}{%
        \begin{tabular}{lcccc}
            \toprule
            \textbf{Model} & \textbf{Tokens} & \textbf{No Pos} & \textbf{Abs Pos} & \textbf{ROPE} \\
            \midrule
            RWKV-X & 10B & 3.08 & 3.10 & 3.11 \\
            \bottomrule
        \end{tabular}
    }
\end{table}

\section{Conclusion}
In this work, we present RWKV-X, a hybrid language model that integrates the efficiency of RWKV for short-range dependencies with a novel sparse attention mechanism for long-range context modeling. By addressing the quadratic complexity limitations of traditional Transformers and the long-context shortcomings of prior linear architectures, RWKV-X achieves linear-time complexity during training and constant-time complexity during inference, enabling scalable processing of sequences up to 1 million tokens.

\section*{Limitations}

While RWKV-X demonstrates strong performance and efficiency in long-context language modeling, several limitations remain. 
First, its sparse attention mechanism, based on top-$k$ chunk selection, is heuristic and may overlook some semantically relevant dependencies.
Second, in our current implementation, sparse attention decoding is slower than that of vanilla RWKV. Further engineering efforts are required to optimize the implementation.


\bibliography{custom}

\newpage
\appendix

\onecolumn

\begin{center}
    {\LARGE \textbf{Supplementary Material for RWKV-X}}\\[2em]
\end{center}

\section{Data and Hyperparameters}
\label{sec:data_Hyperparameters}

\paragraph{Training Data}
RWKV-X training is divided into two phases. 
The first phase, the Alignment Phase, uses the minipile dataset with 1.5 billion tokens. 
The second phase, the Long Context Phase, draws randomly sampled data from the ProLong-64K dataset with a total of 40 billion tokens.

\paragraph{Hyperparameters}
The following hyperparameters were used to train a range of RWKV-X models, from 0.22B to 3.6B parameters, as shown in Table \ref{tab:hyperparameter}.

\begin{table}[h!]
\centering
\resizebox{\textwidth}{!}{
\begin{tabular}{l|cc|cc}
\toprule
\multirow{2}{*}{Hyperparameter} & \multicolumn{2}{c|}{0.22B Model} & \multicolumn{2}{c}{3.6B Model} \\
\cmidrule(lr){2-5}
& Alignment & Long Context & Alignment & Long Context \\
\midrule
Batch size (tokens) & - & 8.192M & 4.096M & 1.024M \\
Context length (tokens) & - & 64,000 & 4,096 & 64,000 \\
Tokens trained (B) & - & 20 & 1.5 & 1 \\
Initial learning rate & - & 1e-5 & 1e-5 & 1e-5 \\
Final learning rate & - & 1e-5 & 1e-5 & 1e-5 \\
Learning rate schedule & - & Constant & Constant & Constant \\
Warmup ratio & - & 0 & 0 & 0 \\
Weight decay & - & 0 & 0 & 0 \\
Optimizer & - & AdamW & AdamW & AdamW \\
DeepSpeed stage & - & 1 & 1 & 1 \\
GPU Configuration & - & 8×H20 & 4×H20 & 8×H200 \\
Total GPU Hours (h) & - & 576 & 6 & 80 \\
\bottomrule
\end{tabular}
}
\caption{Training hyperparameters and compute configurations for RWKV-X models.}
\label{tab:hyperparameter}
\end{table}

\section{Training Efficiency}

\begin{figure}[h!]
    \centering
    \includegraphics[width=0.7\linewidth]{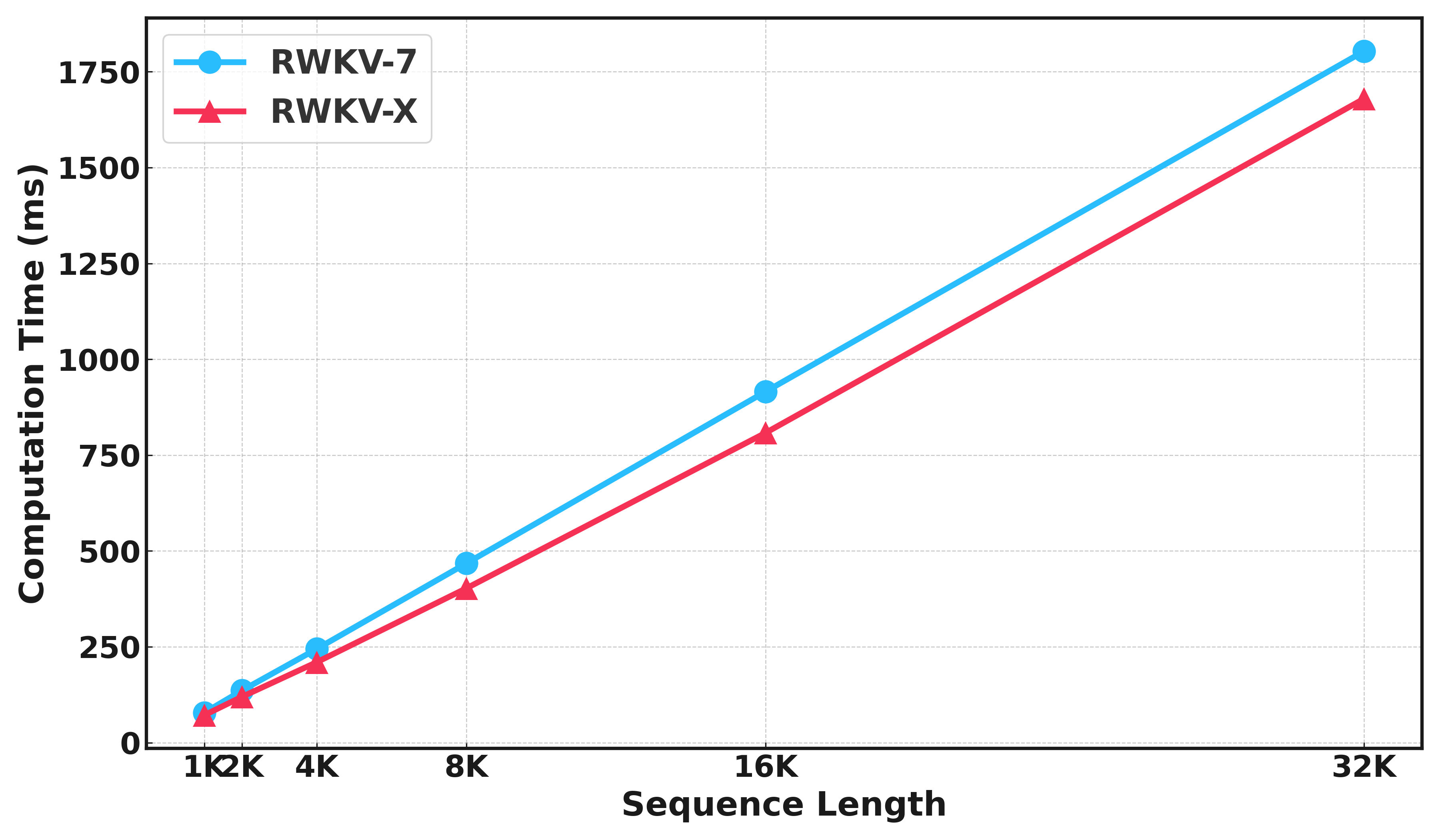}
    \caption{Training efficiency comparison between RWKV-X and RWKV-7}
    \label{fig:train_efficiency}
\end{figure}

Figure~\ref{fig:train_efficiency} illustrates the training efficiency comparison between RWKV-X and RWKV-7 across varying sequence lengths, ranging from 1K to 32K tokens.
As the sequence length increases, both models exhibit approximately linear growth in computation time, consistent with their underlying design. 
Notably, RWKV-X consistently demonstrates lower computation time compared to RWKV-7 at all sequence lengths, highlighting its improved training efficiency. 
The gap in efficiency becomes more pronounced at longer sequence lengths, suggesting that the architectural modifications in RWKV-X more effectively scale with context size. 
These results indicate that RWKV-X offers a more computationally efficient alternative to RWKV-7, particularly for tasks requiring long-context processing.

\section{More on Efficiency Analysis}
\subsection{Comparison of Sparse and Full Attention}
Table~\ref{tab:attention-comparison} presents a comparison between sparse and full attention used in RWKV-X across varying context lengths in terms of latency and memory consumption. Sparse attention exhibits slightly higher prefill latency at shorter context lengths, but shows a clear advantage in decoding latency at larger scales (e.g., 121.99 ms vs. 170.79 ms at 256k context length). Memory usage is nearly identical between the two methods for smaller contexts, but sparse attention maintains a slight efficiency lead as the sequence length increases. Notably, sparse attention provides more consistent decoding performance as context length scales, making it more suitable for long-context applications where decoding speed is critical.

\begin{table}[htbp]
\centering
\caption{Comparison of Sparse and Full Attention on Latency and Memory Usage}
\resizebox{\textwidth}{!}{
\begin{tabular}{lcccc}
\toprule
Context Length & Latency (Prefill) & Memory (Prefill) & Latency (Decoding) & Memory (Decoding) \\
               & Sparse / Full     & Sparse / Full     & Sparse / Full       & Sparse / Full      \\
\midrule
4K     & 517.64 / 511.06    & 8.70 / 8.70     & 41.73 / 41.82     & 8.42 / 8.42 \\
8K     & 643.26 / 660.30    & 9.06 / 9.06     & 39.14 / 34.33     & 8.77 / 8.72 \\
16K    & 1408.31 / 1404.56  & 9.66 / 9.66     & 36.04 / 34.31     & 9.45 / 9.32 \\
32K    & 2960.36 / 2955.05  & 10.96 / 10.96   & 37.03 / 34.39     & 10.81 / 10.69 \\
64K    & 6107.07 / 6103.40  & 13.69 / 13.69   & 38.28 / 41.97     & 13.54 / 13.42 \\
128K   & 12913.58 / 12792.79 & 19.20 / 19.20   & 58.59 / 68.14     & 19.06 / 18.93 \\
256K   & 31668.96 / 31776.53 & 30.17 / 30.17   & 121.99 / 170.79   & 30.02 / 29.90 \\
512K   & 95482.76 / 95824.31 & 52.14 / 52.14   & 289.91 / 323.96   & 51.99 / 51.87 \\
\bottomrule
\end{tabular}
}
\label{tab:attention-comparison}
\end{table}

\section{KV Cache Management for Top-\textit{k} Chunk Sparse Attention}
\label{sec:kv_cache}

In Top-$k$ Chunk Sparse Attention, maintaining a manageable KV cache size is crucial for achieving efficient decoding. 
We adopt a compression strategy to ensure that the past KV cache remains constant in size, regardless of the input sequence length.

\begin{figure}[h!]
    \includegraphics[width=1.5\textwidth]{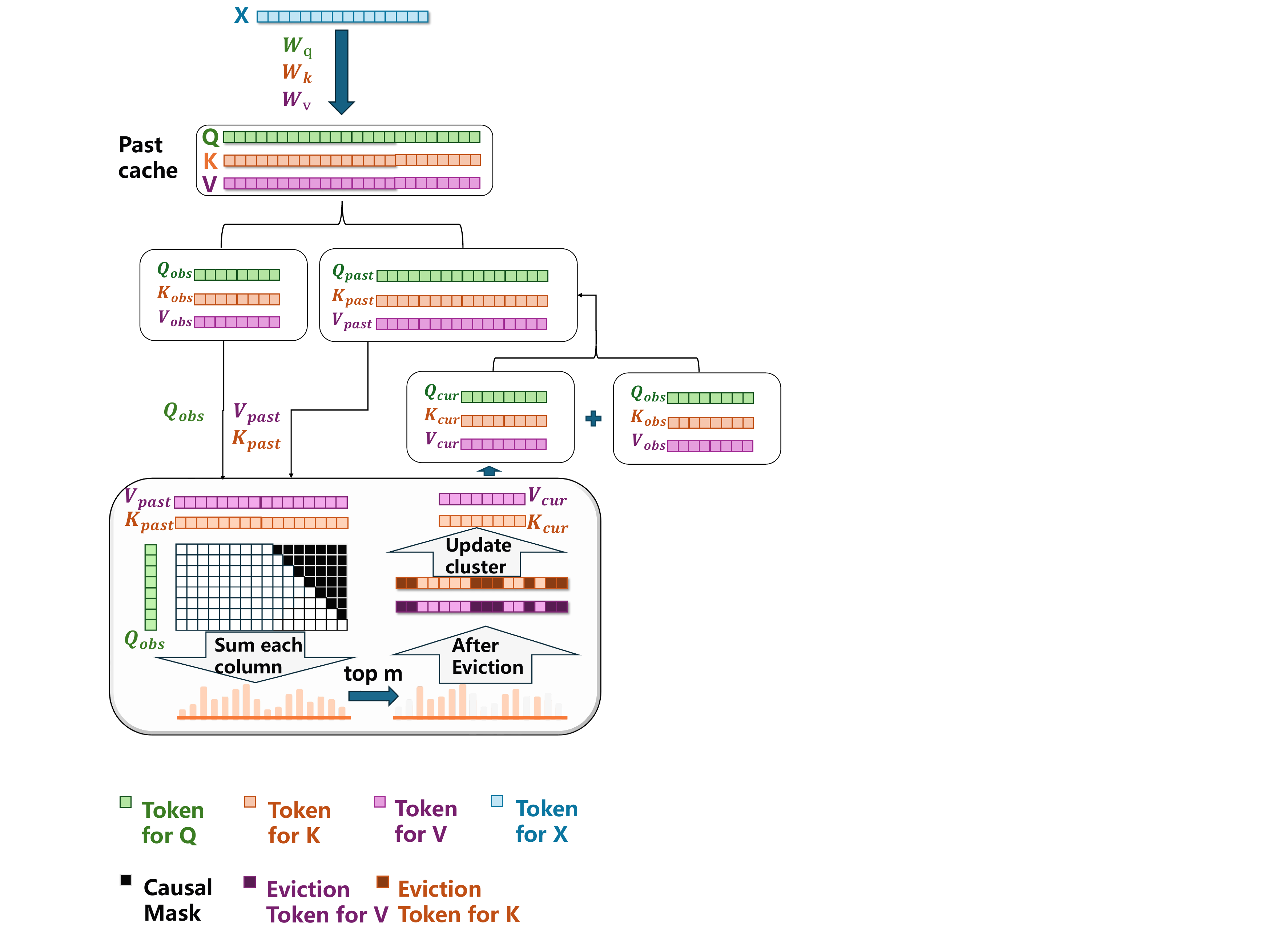}
    \centering
    \caption{Illustration of KV cache management for Top-$k$ Chunk Sparse Attention.}
    \label{fig:app_kv_cache}
\end{figure}

Figure~\ref{fig:app_kv_cache} illustrates the KV cache management process. 
We begin by splitting the past cache into two parts: the \textit{earlier} cached states $(K_{\text{past}}, V_{\text{past}})$ and the \textit{recent} observation window $(K_{\text{obs}}, V_{\text{obs}})$. The observation window contains the most recent tokens, which are always retained due to their relevance to the current context.

To assess the importance of the earlier cached entries, we calculate a cumulative importance score vector by summing the softmax-normalized attention weights over each key.These scores reflect how much past tokens are attended to by the current observation window. Based on this, we retain the top-$m$ entries with the highest importance, where $m$ is a predefined memory budget. The remaining entries are evicted from the cache.

Following eviction, we update the cache by concatenating the selected top-$m$ keys and values with those from the observation window $(K_{\text{obs}}, V_{\text{obs}})$, producing a compressed cache that preserves essential information while capping memory usage.

Specifically, we dynamically select the most relevant cached entries based on their cumulative attention scores with respect to the observation window queries, and discard less important entries. 
This selective compression significantly reduces memory footprint during long-sequence generation while preserving essential contextual information for accurate predictions.

\end{document}